# FONTNET: ON-DEVICE FONT UNDERSTANDING AND PREDICTION PIPELINE

*Rakshith S, Rishabh Khurana, Vibhav Agarwal, Jayesh Rajkumar Vachhani, Guggilla Bhanodai*

Samsung R&D Institute India, Bangalore, India – 560037

## ABSTRACT

Fonts are one of the most basic and core design concepts. Numerous use cases can benefit from an in depth understanding of Fonts such as Text Customization which can change text in an image while maintaining the Font attributes like style, color, size. Currently, Text recognition solutions can group recognized text based on line breaks or paragraph breaks, if the Font attributes are known multiple text blocks can be combined based on context in a meaningful manner. In this paper, we propose two engines: Font Detection Engine, which identifies the font style, color and size attributes of text in an image and a Font Prediction Engine, which predicts similar fonts for a query font. Major contributions of this paper are three-fold: First, we developed a novel CNN architecture for identifying font style of text in images. Second, we designed a novel algorithm for predicting similar fonts for a given query font. Third, we have optimized and deployed the entire engine On-Device which ensures privacy and improves latency in real time applications such as instant messaging. We achieve a worst case On-Device inference time of 30ms and a model size of 4.5MB for both the engines.

## 1. INTRODUCTION

In this modern digital age, approximately 3.2 billion people are accessing social networking platforms through their mobile phones. The use of visual media such as memes, stickers and GIFs is ever increasing and the current options are limited. There is an urgent need for customizable stickers and GIFs which can be personalized. By identifying the font style, color and size, users can reuse any one template for multiple scenarios. The options available can be further increased by providing a list of aesthetically similar fonts. Such a subtle art of selecting visually similar fonts requires a sophisticated skillset. Also the font names themselves are rarely meaningful which makes this task really challenging for most users.

Artificial intelligence is moving towards an On-Device platform from the cloud platform for better reliability, more privacy, and consistent performance. However, it requires lightweight, fast, and accurate neural network models to run on a resource constraint mobile platform. This is another crucial aspect we would like to focus in this paper.

Even though there are websites that detect font styles from images (WhatTheFont, Font Matcherator, etc.), there exists no solution which works offline. Similar limitations exist for font color and font size as well. Websites such as "Identifont", suggest similar fonts based on proximity between visual features with respect to a query font, but there is a lack of variety in these fonts.

In this paper, we propose a Font Detection Engine (FDE), which is capable of identifying all the font attributes (style, color, size) of text in images. By observing the current literature we find that there have been many attempts to solve this task. For example, Wang et al. [1] have used a CNN based architecture for font classification. Their work attempts to solve the Visual Font Recognition (VFR) task. Even though their network accommodates both real world as well as synthetic data and supports a larger set of fonts, their network is infeasible to be used On-Device because of a large model size and inference time. In [2] the authors use a similar approach to include both synthetic and real world data, but they have implemented it with VGG [9] and AlexNet [8] architectures and have the same shortcomings as [1]. Chen et al. [3] use a Nearest Class Mean Classifier based approach for solving the VFR task. Such a statistical approach, makes their model scalable and accommodates new font classes. Since it is based on a local feature embedding of the image, it is not very accurate for the font detection task. Furthermore, it is very difficult to define such local feature vectors which suits all font types. The results for wild images prove that this approach does not work well for noisy images with complex backgrounds. In [4], Yifan Chang has also used a CNN for identifying typefaces of Chinese text in images. It exhibits good results only for black and white synthetic images.

There have been multiple attempts for predicting suitable fonts for design tasks. In [15], the authors adopt a deep neural network based model to predict most suitable fonts for a given web design. They use visual features from a CNN along with semantic tags from the webpage for understanding context. In [11], the authors follow a complicated method to learn the distance/similarity between two fonts for predicting visually similar fonts. They use crowd sourced data which is costly and cumbersome to collect. Therefore, we use an alternate approach for finding similar fonts as discussed in section 2.

## 2. PROPOSED METHOD

### 2.1. Font Detection Engine

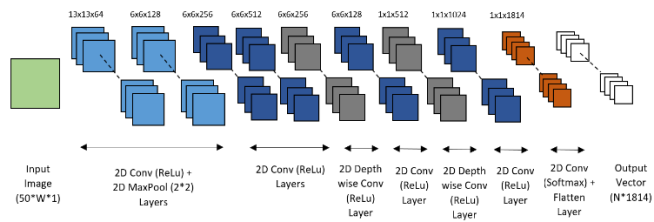

**Fig. 1.** The Fully Convolutional Neural Network architecture for Font style detection.

FDE uses localized text regions generated by a text recognition API (Google ML Kit) [16]. Our engine captures all the font attributes (style, color, size) of the text. We use deep CNN architecture for the font style identification. For training the model we use a dataset of 1814 fonts (Google Fonts [13]) with 700 images each. The dataset consists of 3 image resolutions, 16 text sizes, 8 text colors, lower/upper case, and augmentations like scaling, contrast etc.

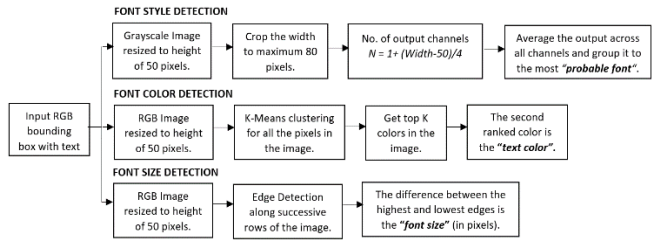

**Fig. 2.** FDE procedure. Here, 'Width' is min (80, rescaled width). A factor of 4 in the denominator is due the 2 MaxPool layers.

#### 2.1.1. Deep CNN
We start with the architecture from the DeepFont paper [1] with a fixed input size of 105x105 comprising of Convolutional, Normalization and MaxPool layers, followed by 3 Dense layers with Softmax activation for the final classification. The 3 Dense layers made the model size excessively large and exhibited poor validation accuracy due to overfitting. Therefore, we reduced the number of Dense layers to 1. Though this improved the accuracy, the model size and inference time were still high. The fixed input size, requires preprocessing operations (dividing the input image into individual patches of size 105x105) before inferring the model which is another limitation.

Due to these shortcomings, we introduced necessary changes as depicted in Figure 1. To solve the fixed input problem, we decided to use a Fully Convolutional Network (FCN) architecture with a sliding window approach. To reduce the model size, we decreased the input height from 105 pixels to 50. The final convolutional layer uses a Softmax activation function and outputs the probabilities for each font. There are no Dense layers and the number of output channels is equal to the effective number of 50x50 patches created due to the stride of the sliding window. This stride is fixed by the number of MaxPool layers. As this stride value increases the inference time of the model reduces. From Figure 2, it is evident that each RGB image is converted to grayscale. The image is also uniformly rescaled to a height of 50 pixels and a width less than 80 pixels to limit the maximum inference time.

#### 2.1.2. Text Color
For identifying the text color we used an algorithm inspired from K-Means clustering. We clustered the colors in an image based on their individual RGB pixel values and picked the top 'K' colors. We consider the pixel values of the three channels as points in 3D space. Euclidean distance was used as a metric to assign each point to one of the K clusters. The centroids of all the K clusters are updated accordingly.

$$d = \sqrt{[(R_1 - R_2)^2 + (G_1 - G_2)^2 + (B_1 - B_2)^2]} \qquad (1)$$

Here $(R_1, G_1, B_1)$ is $color_1$ and $(R_2, G_2, B_2)$ is $color_2$ and $d$ is the distance between them. Once the centroid locations are constant after successive iterations we choose the top K colors as the K centroids that were obtained and rank them based on the number of points in the cluster (i.e. the area they occupy in the image). Based on our analysis over a wide variety of images, we found that the text color always occupies the second highest area in the bounding box.

#### 2.1.3. Text Size
Using a simple edge detection algorithm, we find the first and last vertical edge in an image. The difference between them is chosen as the maximum height of the text. In the algorithm, we include a threshold parameter $T$. This parameter helps in differentiating between stray edges due to the background noise and the edges of text in the image. Here $I$ represents the $R$, $G$ and $B$ channels. An edge exists in the $i^{th}$ row of the image, only if this condition is satisfied.

$$abs(I[i,j] - I[i+1,j]) > T \quad (2)$$

**2.2. Font Prediction Engine**

This engine predicts a list of visually similar fonts for any given query font from our dataset of 1762 fonts. These similar fonts can be used to replace the query font without compromising on the overall aesthetic quality of the image.

*2.2.1. Dataset Creation*
For estimating the similarity of two fonts, we need a mapping between a font and a set of attributes. These attributes are basically adjectives that explain the visual characteristics of the font, example 'legible', 'serif', 'thin' etc. There are 37 such attributes and their range is [0,100]. In [11], they use crowdsourced data to obtain such a mapping. This crowdsourced dataset is available only for 200 fonts from the Google Fonts [13] website.

**Fig. 3.** Basis for selecting high priority attributes

In order to extend this dataset, we use a kNN based algorithm [10], where we consider the $k$ nearest neighbors of a new font to calculate its attribute vector $\vec{f}$. In [10], the authors have proven the effectiveness of using such an approach for extending the dataset. For finding the neighbors we use 200-dimensional CNN embeddings $\vec{g}$ defined for 1883 Google fonts [14] from a pre-trained network. Out of the 1883 font dataset and 200 font dataset, there are 156 common fonts. These 156 fonts will be used as seed data for finding the nearest neighbors. Out of the 1883 fonts, only 1762 fonts are supported for Android devices, therefore we can extend our dataset up to 1762 fonts (156 old fonts + 1606 new fonts).

$$w_i = \frac{1}{k-1}\left(\frac{\sum_{\substack{j=1 \\ i \neq j}}^{k} d(\vec{g},\vec{g_j})}{\sum_{j=1}^{k} d(\vec{g},\vec{g_j})}\right) \quad (3)$$

$$\vec{f} = \sum_{i=1}^{k} w_i \vec{f_i} \quad (4)$$

Once we get the nearest neighbors for each new font, we then calculate its attribute vector using the weighted average of attribute vectors of these nearest neighbors as shown in (3) and (4). Here, $d(\vec{g_1},\vec{g_2})$ is the L2-distance between the two embeddings, this metric is used to find the nearest neighbors. The same procedure is repeated for all the 1606 fonts.

*2.2.2. Font Prediction Algorithm*
This algorithm predicts visually similar fonts to a given query font using the attribute vector. We first select 11 most *faithful* attributes, i.e. the attributes whose values and their manifestation in the visual properties of the fonts is in congruence. We define these as *priority* attributes '$p$' as depicted in Figure 4. In addition to this, we define relative weights for each of these priority attributes using weight vector '$w$'. We choose an Interval vector '$I$' that defines ranges for all attributes which behave as search windows around the attributes of the query font. We consider fonts which lie in this range and then rank them based on their weighted distance with respect to the priority attributes.

In Figure 3, we have shown the top-4 fonts for each of the attributes. In (a), the attributes (italic, thin) describe the visual nature of the fonts and therefore we consider them in our list of priority attributes. But in (b), we notice that the attributes (cursive, delicate) and the fonts have no correlation with each other. Therefore these are less reliable and we don't consider them in the Font Prediction algorithm.

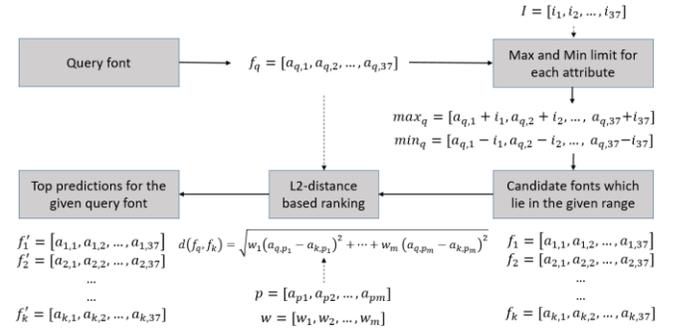

**Fig. 4.** Overview of the Font Prediction algorithm

Figure 4 outlines the high level description of the font prediction algorithm. We start with our query font $f_q$. From our newly created dataset we get its corresponding 37-dimensional attribute vector. The interval vector $I$ defines the search space for nearby fonts. From $f_q$ and $I$ we get the search window for each attribute. While searching for the prospective candidates we only consider the priority attributes $p$. Once we have the list of candidate fonts, we use a weighted Euclidean distance measure to rank the candidate fonts. This list of sorted fonts is our final list of predicted fonts.

**2.3. Model Compression and On-Device Deployment**

A very crucial part of this paper is to deploy both the engines on a mobile device and use it in real time. We applied multiple optimizations to our model making it more efficient with respect to time and space constraints. Firstly, we reduced the input height from 105 to 50 while preserving the aspect ratio. This reduced the model size drastically while maintaining its accuracy. We used "Depth-wise Separable Convolutional" layers which reduce the model size and work well in edge devices as shown in [12]. This is the final model (shown in Figure 1) used in the FDE which shows a 70% reduction in model size (15MB to 4.5 MB) and 86% reduction in inference time (300ms to 40 ms). This model was converted to TFLite and used in an Android application. We used native inferencing using C++ to get the optimal results.

## 3. OBSERVATIONS AND RESULTS

### 3.1. Font Detection Engine

*3.1.1. Model Comparison*
From Table 1, it is evident that there is a drastic reduction in the number of parameters in the FCN compared to the model with a Dense layer and also improves the validation accuracy. For the inference time calculation, we use two sets of images, one with a width of 50 pixels and the other with 80 pixels. For the inference of the dense layer model, we created image patches of size 50x50 after a width of every 4 pixels. These are the same patches that are being seen internally by the FCN.

| Model | Dataset | Size (MB) | Model parameters | Inference time (ms) | | Validation Accuracy (Top-1) |
|---|---|---|---|---|---|---|
| | | | | 50px | 80px | |
| Dense Model | Google Fonts | 17.8 | 18.2 M | 48 | 115 | 0.783 |
| FCN | Google Fonts | 4.5 | 4.6 M | 12 | 29 | 0.885 |
| FCN | VFR Dataset | 5.0 | 5.1 M | 14 | 33 | 0.782 |

**Table. 1.** Comparison between models. (All results were evaluated on a mobile device with 12 GB RAM and an Octa-core processor).

With FCN we achieved a worst case inference time of 30ms. This is one of the crucial improvements which made it possible to port the model on a mobile device and achieve seamless experience. We achieved a Top-1 validation accuracy of 78% on the VFR synthetic dataset with a model size as small as 5MB. In [1], we see that the DeepFont model has a Top-1 accuracy of 80% on the VFR synthetic dataset. But their model has a whopping 26M parameters which is 5 times larger than our model. The VFR model is also compressed using matrix factorization techniques. Therefore in comparison to the state of the art, our model has a commendable accuracy considering the massive difference in model size.

*3.1.2. Font Detection in Full Sized Digital Images*
Figure 5 shows detailed steps of the FDE pipeline for replacing the original message ("Happy Birthday") with a new message ("Happy New Year") while preserving the background. The FDE detects style, color and size on the text region selected by the user. Inpainting API from Open CV [17] is used for removing original text. The detected font attributes are applied on the new input message and replaced in the image. The entire process takes less than 30ms.

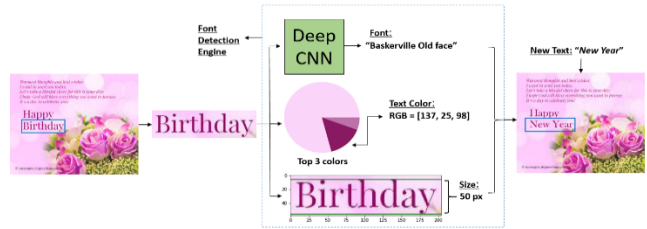

**Fig. 5.** FDE pipeline for a sample image.

### 3.2. Font Prediction Engine

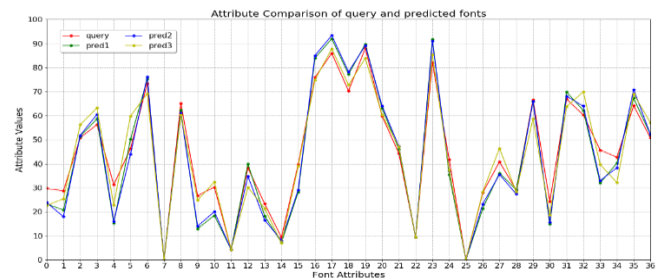

**Fig. 6.** Font Prediction results for three different query fonts

The results for the FPE can be seen in Figure 6. For every query font, we have listed the top 3 predictions. We notice that the predicted fonts have many common qualities with the query font. For a cursive font such as *Sacramento* all of its top predictions also have the same nature and similar stroke thickness. For a thick and bold font such as *Erica One*, the predictions have a consistent visual quality. The serif quality is preserved for a font such as *Abril Fatface*. The predicted fonts have the same features and preserves the underlying intent as much as possible while also maintaining the requisite amount of diversity. The features and styles to be preserved is controlled by tuning the priority attributes and the weight vector. In [11], the authors have used a learned "distance metric" over crowd sourced data to quantify the similarity between two fonts. Compared to this, we have devised a simpler method.

**Fig. 7.** Attribute Comparison between Query font (Sacramento) and Predicted fonts.

Figure 7 depicts the 37 font attributes of the font Abril Fatface and its top three predictions. The predicted fonts closely resemble the query font in terms of the distance between attributes. This is reflected in their visual characteristics in Figure 6. The attributes showing large deviation in values (eg: attribute 9, 10, 17 etc.) belong to the low priority attributes which are not being used by the FPE.

## 4. CONCLUSIONS AND FUTURE WORK

In this paper, we implemented a Font Detection Engine, which identifies font style, font color and font size of text in an image. We developed a novel algorithm for predicting visually similar fonts for a given query font. Both the engines were then deployed on a mobile device and can work in real time. This Font Detection Engine can be used in applications such as, text customization and recognition. Both engines reduce the design effort required for selecting suitable fonts. As future work, we would want to make our current neural network scalable, to include extra fonts and also fonts from languages other than English. We can also try to link fonts and the corresponding moods they convey, making the prediction process much more robust and accurate.